\documentclass[acmsmall]{acmart}

\AtBeginDocument{%
  }

\setcopyright{acmlicensed}
\copyrightyear{2024}
\acmYear{2024}
\acmDOI{XXXXXXX.XXXXXXX}

\acmConference[Conference acronym 'XX]{Make sure to enter the correct
  conference title from your rights confirmation emai}{June 03--05,
  2024}{Woodstock, NY}
\acmISBN{978-1-4503-XXXX-X/18/06}

\begin{document}

\title{Leveraging Large Language Models for Entity Matching}

\author{Qianyu Huang}
\email{qianyuhuang19@gmail.com
}
\affiliation{%
  \institution{National Taiwan University of Science \& Technology}
  \city{Taipei}
  \country{Taiwan}
}

\author{Tongfang Zhao}
\affiliation{%
  \institution{National Taiwan University of Science \& Technology}
  \city{Taipei}
  \country{Taiwan}}

\renewcommand{\shortauthors}{Trovato et al.}

\begin{abstract}
Entity matching (EM) is a critical task in data integration, aiming to identify records across different datasets that refer to the same real-world entities. Traditional methods often rely on manually engineered features and rule-based systems, which struggle with diverse and unstructured data. The emergence of Large Language Models (LLMs) such as GPT-4 offers transformative potential for EM, leveraging their advanced semantic understanding and contextual capabilities. This vision paper explores the application of LLMs to EM, discussing their advantages, challenges, and future research directions. Additionally, we review related work on applying weak supervision and unsupervised approaches to EM, highlighting how LLMs can enhance these methods.
\end{abstract}




\maketitle

\section{Introduction}
Entity matching (EM) is a foundational task in data integration, essential for applications in various domains including data warehousing, e-commerce, healthcare, and social media~\cite{li2021deep, barlaug2021neural, wu2020zeroer, yan2020entity, mudgal2018deep}. The objective of EM is to identify and link records that refer to the same real-world entities across disparate data sources. Traditional EM techniques, which often rely on rule-based systems or machine learning models with manually engineered features, face significant challenges when dealing with diverse, noisy, and unstructured data.

Recent advancements in natural language processing (NLP) have led to the development of Large Language Models (LLMs) like GPT-4, which are trained on vast corpora of text and possess advanced capabilities in understanding and generating human-like text~\cite{kasneci2023chatgpt}. These models have shown remarkable performance in various NLP tasks, suggesting their potential to revolutionize EM by addressing the limitations of traditional methods~\cite{peeters2023entity}.

This paper envisions a future where LLMs are central to EM, offering improved accuracy, scalability, and adaptability. We will explore the advantages of LLMs in EM, discuss related work on weak supervision and unsupervised EM, outline the challenges, and propose future research directions.

\section{Advantages of LLMs in Entity Matching}

\subsection{Semantic Understanding}
One of the most significant advantages of LLMs is their ability to understand the semantics and context of entities~\cite{chang2024survey,zheng2023adapting}. Traditional EM methods often rely on syntactic similarity measures, which can be inadequate when dealing with variations in how entities are described. For instance, consider the following records:

\begin{itemize}
    \item Record 1: "Microsoft Corporation"
    \item Record 2: "MSFT"
\end{itemize}

A traditional EM system might struggle to recognize that these records refer to the same entity. In contrast, an LLM can leverage its extensive training data to understand that "Microsoft Corporation" and "MSFT" are semantically equivalent, thus improving matching accuracy.

\subsection{Minimal Feature Engineering}
Traditional EM methods require extensive feature engineering to capture relevant attributes for matching, which is both time-consuming and domain-specific. LLMs, on the other hand, can leverage their pre-trained knowledge to generate contextual embeddings for entities, reducing the need for manual feature engineering. This capability allows LLMs to adapt to various domains with minimal effort, making them highly versatile.

\subsection{Adaptability}
LLMs can be fine-tuned on specific datasets to capture domain-specific nuances~\cite{li2021quantifying, rao2024normad}. This adaptability is crucial for EM tasks, where the nature of data can vary significantly across different applications. Fine-tuning allows LLMs to learn from a small amount of labeled data, improving their performance on specific EM tasks without extensive re-engineering.

\subsection{Handling Unstructured Data}
Many real-world datasets contain unstructured text, making it challenging for traditional EM methods to process and match entities accurately. LLMs excel at understanding unstructured text, enabling them to perform EM tasks involving diverse data formats~\cite{tan2023unstructured, liu2022multimodal}. This capability is particularly valuable in domains like social media and e-commerce, where data is often noisy and unstructured.

\section{Related Work}

\subsection{Unsupervised Entity Matching}
Unsupervised EM methods aim to match entities without relying on labeled training data. These methods often involve clustering techniques, similarity measures, and iterative refinement to identify matching records~\cite{zhang2020unsupervised}. While unsupervised methods are valuable in scenarios where labeled data is scarce, they often struggle with high-dimensional and noisy data~\cite{ge2021collaborem}.

LLMs can significantly enhance unsupervised EM by providing rich, contextual embeddings for entities. These embeddings capture semantic relationships and can be used to compute more accurate similarity measures. Additionally, LLMs can generate synthetic data or augment existing data, improving the robustness of unsupervised methods.

Consider an unsupervised EM task involving matching research papers across different academic databases. Traditional methods might rely on string similarity measures for titles and author names, which can be insufficient due to variations in naming conventions and abbreviations. By leveraging LLMs, we can generate embeddings that capture the semantic content of paper titles and author names, improving the accuracy of similarity measures and clustering algorithms used in unsupervised EM.

\subsection{Use of Pre-trained Models in Entity Matching}
Pre-trained models have become increasingly popular in the field of entity matching due to their ability to leverage vast amounts of data and transfer learning capabilities~\cite{li2020deep}. Models like BERT, RoBERTa, and GPT have been fine-tuned for specific EM tasks~\cite{li2023effective}, demonstrating significant improvements in accuracy and robustness. These models are trained on extensive text corpora, capturing a wide range of linguistic patterns and contextual nuances that are invaluable for EM. For example, BERT-based models have been employed to generate contextual embeddings for entities, which are then used to compute similarity scores and identify matches. The pre-trained nature of these models allows them to be quickly adapted to new domains with minimal additional training, making them highly versatile and efficient for various EM applications.

\subsection{Weak Supervision in Entity Matching}
Weak supervision is an approach that leverages noisy, limited, or imprecise sources of supervision to train machine learning models~\cite{ratner2017snorkel}. In the context of EM, weak supervision can involve using heuristics, domain knowledge, or external databases to generate training data~\cite{wu2023ground}. This approach addresses the challenge of obtaining large amounts of labeled data, which is often a bottleneck in training supervised EM models.

Recent work in weak supervision for EM includes methods like Panda~\cite{wu2021demonstration}, which uses multiple weak supervision sources to generate probabilistic labels. These methods have shown promise in improving EM performance by leveraging diverse sources of information. LLMs can enhance weak supervision by providing high-quality embeddings and contextual information, improving the accuracy of weakly supervised models.

For example, consider a scenario where we have limited labeled data for matching product names across different e-commerce platforms. Traditional weak supervision methods might use heuristics based on string similarity or external databases to generate training data. By incorporating LLMs, we can generate more accurate embeddings for product names, capturing semantic similarities that traditional methods might miss. This integration can lead to improved training data quality and, consequently, better EM performance.

\subsection{Active Learning in Entity Matching}
Active learning is a powerful technique for improving the performance of entity matching models by iteratively selecting the most informative examples for labeling. This approach is particularly useful in scenarios where labeled data is scarce or expensive to obtain. In the context of EM, active learning algorithms identify uncertain or ambiguous matches and query human experts for labels, thereby refining the model's accuracy with each iteration~\cite{meduri2020comprehensive}. 
Techniques such as uncertainty sampling, where the model selects examples it is least confident about, and diversity sampling, where the model selects a diverse set of examples, have been effectively applied to EM tasks~\cite{brunner2019entity}. By focusing labeling efforts on the most challenging cases, active learning not only enhances model performance but also reduces the overall labeling cost, making it a practical solution for real-world EM applications.

\section{Challenges and Solutions}

\subsection{Scalability}
LLMs are computationally intensive, requiring significant resources for both training and inference. This poses a challenge for large-scale EM tasks, where efficiency is crucial. To address scalability, we propose the following solutions:

\begin{enumerate}
    \item \textbf{Model Distillation}: Model distillation involves training a smaller, more efficient model to mimic the behavior of a larger LLM~\cite{wu2024divide, peng2024metaie}. This approach can significantly reduce computational requirements while maintaining high performance.
    \item \textbf{Efficient Indexing}: Developing efficient indexing methods for LLM-generated embeddings can speed up similarity computation~\cite{hua2023index}. Techniques such as approximate nearest neighbor search can be employed to handle large datasets.
    \item \textbf{Distributed Computing}: Leveraging distributed computing frameworks can help manage the computational load of LLMs, enabling parallel processing of large datasets~\cite{huang2024fast}.
\end{enumerate}

\subsection{Data Privacy}
Handling sensitive data in EM tasks requires robust privacy-preserving techniques. LLMs, trained on large corpora, may inadvertently expose sensitive information. To mitigate privacy concerns, we propose the following approaches:

\begin{enumerate}
    \item \textbf{Differential Privacy}: Implementing differential privacy techniques can help protect individual data points while allowing aggregate analysis. This approach ensures that the output of the LLM does not reveal specific details about any single record~\cite{behnia2022ew}.
    \item \textbf{Federated Learning}: Federated learning enables training models across decentralized devices while keeping data local. This approach can be used to train LLMs on sensitive data without transferring it to a central server, preserving privacy~\cite{kuang2023federatedscope, fan2023fate}.
\end{enumerate}

\subsection{Domain Adaptation}
Ensuring that LLMs generalize well across different domains remains a challenge. While fine-tuning can improve performance on specific datasets, it may not always be sufficient for diverse EM tasks. To enhance domain adaptation, we propose the following strategies:

\begin{enumerate}
    \item \textbf{Transfer Learning}: Transfer learning involves pre-training LLMs on a large, diverse corpus and then fine-tuning them on domain-specific data. This approach can help LLMs retain general knowledge while adapting to specific domains~\cite{chronopoulou2019embarrassingly}.
\item \textbf{Domain Adaptation Techniques}: Techniques such as domain adversarial training and multi-task learning can help LLMs generalize across different domains~\cite{saad2023udapdr}. These methods involve training the model to perform well on both source and target domains, improving robustness.
\end{enumerate}

\subsection{Interpretability}
LLMs are often seen as black boxes, making it challenging to understand their decision-making process. Interpretability is crucial for gaining user trust and improving model debugging. To address this challenge, we propose the following solutions:

\begin{enumerate}
    \item \textbf{Explainable AI (XAI) Techniques}: Developing XAI techniques for LLMs can help make their decisions more transparent. Methods such as attention visualization and feature importance analysis can provide insights into how LLMs make matching decisions~\cite{yu2023temporal, wu2024usable}.
    \item \textbf{Human-in-the-Loop Systems}: Incorporating human feedback into the EM process can improve interpretability~\cite{zeng2024similar}. By allowing users to review and validate matches, we can gain insights into the model's behavior and identify areas for improvement.
\end{enumerate}

\section{Future Research Directions}

\subsection{Hybrid Models}
Combining LLMs with traditional EM methods can leverage the strengths of both approaches. Hybrid models can use LLMs for generating high-quality embeddings and contextual information, while traditional methods can provide domain-specific rules and heuristics. This integration can enhance overall performance and robustness.

For instance, a hybrid model for matching customer records across different databases might use LLMs to generate embeddings for customer names and addresses, capturing semantic similarities. Traditional rule-based methods can then be applied to specific attributes like phone numbers and email addresses, ensuring accurate matching.

\subsection{Interactive EM Systems}
Developing interactive EM systems where human experts can provide feedback to LLMs can improve accuracy iteratively. These systems can leverage active learning techniques, where the model queries the user for labels on uncertain matches, gradually refining its performance.

Consider an interactive EM system for matching medical records across different hospitals. The system can present uncertain matches to domain experts, who can provide feedback on whether the records refer to the same patient. This feedback can be used to fine-tune the LLM, improving its accuracy over time.

\subsection{Cross-lingual EM}
Extending LLM-based EM to handle multilingual datasets can enable global data integration. Cross-lingual EM involves matching entities across different languages, which is challenging due to variations in naming conventions and linguistic structures.

LLMs, trained on multilingual corpora, can generate embeddings that capture cross-lingual semantic relationships. These embeddings can be used to compute similarity measures for entities in different languages, facilitating accurate matching. For example, matching product descriptions across e-commerce platforms in different countries can benefit from cross-lingual embeddings generated by LLMs.

\subsection{Real-time EM}
Exploring methods to perform entity matching in real-time is crucial for applications in e-commerce, social media, and other dynamic environments. Real-time EM requires efficient algorithms that can process and match entities as new data arrives.

LLMs can be integrated with streaming data processing frameworks to enable real-time EM. Techniques such as incremental learning, where the model is updated continuously with new data, can help maintain high performance in dynamic environments. For instance, a real-time EM system for social media platforms can match user profiles and posts as they are created, ensuring accurate and up-to-date entity linkage.

\section{Conclusion}
The application of LLMs to entity matching presents a transformative opportunity to improve accuracy, scalability, and adaptability in data integration tasks. By leveraging their semantic understanding and contextual capabilities, LLMs can address many of the limitations of traditional EM methods. However, challenges such as scalability, data privacy, domain adaptation, and interpretability need to be addressed to fully realize their potential. Continued research and innovation in this area promise to unlock new possibilities for efficient and effective entity matching.

In summary, LLMs offer a promising new approach to EM, with the potential to revolutionize data integration tasks across various domains. By exploring hybrid models, interactive systems, cross-lingual EM, and real-time EM, we can further enhance the capabilities of LLMs and address the challenges they face. As we continue to advance our understanding and application of LLMs in EM, we move closer to realizing their full potential in transforming data integration and beyond.

\bibliographystyle{ACM-Reference-Format}
\bibliography{sample-base}

\end{document}